\documentclass[12pt]{article}          
\usepackage{amsmath}
\usepackage{algorithm}
\usepackage{algorithmic}
\usepackage{fancyhdr}
\usepackage{graphicx}

\newcommand\balpha{\ensuremath{\mbox{\boldmath$\alpha$}}}
\newcommand\bbeta{\ensuremath{\mbox{\boldmath$\beta$}}}
\newcommand\btheta{\ensuremath{\mbox{\boldmath$\theta$}}}
\newcommand\bphi{\ensuremath{\mbox{\boldmath$\phi$}}}
\newcommand\bpi{\ensuremath{\mbox{\boldmath$\pi$}}}
\newcommand\bpsi{\ensuremath{\mbox{\boldmath$\psi$}}}
\newcommand\bmu{\ensuremath{\mbox{\boldmath$\mu$}}}
\newcommand\fig[1]{\begin{center} \includegraphics{#1} \end{center}}
\newcommand\Fig[4]{\begin{figure}[ht] \begin{center} \includegraphics[scale=#2]{#1} \end{center} \caption{\label{fig:#3} #4} \end{figure}}
\newcommand\FigTop[4]{\begin{figure}[t] \begin{center} \includegraphics[scale=#2]{#1} \end{center} \caption{\label{fig:#3} #4} \end{figure}}
\newcommand\FigStar[4]{\begin{figure*}[ht] \begin{center} \includegraphics[scale=#2]{#1} \end{center} \caption{\label{fig:#3} #4} \end{figure*}}
\newcommand\aside[1]{\quad\text{[#1]}}
\newcommand\interpret[1]{\llbracket #1 \rrbracket} \DeclareMathOperator*{\tr}{tr}
\DeclareMathOperator*{\sign}{sign}
\newcommand\citet\newcite
\newcommand\citep\cite

\DeclareMathOperator*{\var}{var} \DeclareMathOperator*{\cov}{cov} \DeclareMathOperator*{\diag}{diag} \newcommand\p[1]{\ensuremath{\left( #1 \right)}} \newcommand\pa[1]{\ensuremath{\left\langle #1 \right\rangle}} \newcommand\pb[1]{\ensuremath{\left[ #1 \right]}} \newcommand\pc[1]{\ensuremath{\left\{ #1 \right\}}} \newcommand\eval[2]{\ensuremath{\left. #1 \right|_{#2}}} \newcommand\inv[1]{\ensuremath{\frac{1}{#1}}}
\newcommand\half{\ensuremath{\frac{1}{2}}}
\newcommand\R{\ensuremath{\mathbb{R}}} \newcommand\Z{\ensuremath{\mathbb{Z}}} \newcommand\inner[2]{\ensuremath{\left< #1, #2 \right>}} \newcommand\mat[2]{\ensuremath{\left(\begin{array}{#1}#2\end{array}\right)}} \newcommand\eqn[1]{\begin{align} #1 \end{align}} \newcommand\eqnl[2]{\begin{align} \label{eqn:#1} #2 \end{align}} \newcommand\eqdef{\ensuremath{\stackrel{\rm def}{=}}} \newcommand{\1}{\mathbb{I}} \newcommand{\bone}{\mathbf{1}} \newcommand{\bzero}{\mathbf{0}} \newcommand\refeqn[1]{(\ref{eqn:#1})}
\newcommand\refeqns[2]{(\ref{eqn:#1}) and (\ref{eqn:#2})}
\newcommand\refchp[1]{Chapter~\ref{chp:#1}}
\newcommand\refsec[1]{Section~\ref{sec:#1}}
\newcommand\refsubsec[1]{Section~\ref{subsec:#1}}
\newcommand\refsecs[2]{Sections~\ref{sec:#1} and~\ref{sec:#2}}
\newcommand\reffig[1]{Figure~\ref{fig:#1}}
\newcommand\reffigs[2]{Figures~\ref{fig:#1} and~\ref{fig:#2}}
\newcommand\reffigss[3]{Figures~\ref{fig:#1},~\ref{fig:#2}, and~\ref{fig:#3}}
\newcommand\reffigsss[4]{Figures~\ref{fig:#1},~\ref{fig:#2},~\ref{fig:#3}, and~\ref{fig:#4}}
\newcommand\reftab[1]{Table~\ref{tab:#1}}
\newcommand\refapp[1]{Appendix~\ref{sec:#1}}
\newcommand\refthm[1]{Theorem~\ref{thm:#1}}
\newcommand\refthms[2]{Theorems~\ref{thm:#1} and~\ref{thm:#2}}
\newcommand\reflem[1]{Lemma~\ref{lem:#1}}
\newcommand\reflems[2]{Lemmas~\ref{lem:#1} and~\ref{lem:#2}}
\newcommand\refalg[1]{Algorithm~\ref{alg:#1}}
\newcommand\refalgs[2]{Algorithms~\ref{alg:#1} and~\ref{alg:#2}}
\newcommand\refex[1]{Example~\ref{ex:#1}}
\newcommand\refexs[2]{Examples~\ref{ex:#1} and~\ref{ex:#2}}
\newcommand\refprop[1]{Proposition~\ref{prop:#1}}
\newcommand\refdef[1]{Definition~\ref{def:#1}}
\newcommand\refcor[1]{Corollary~\ref{cor:#1}}
\newcommand\Chapter[2]{\chapter{#2}\label{chp:#1}}
\newcommand\Section[2]{\section{#2}\label{sec:#1}}
\newcommand\Subsection[2]{\subsection{#2}\label{sec:#1}}
\newcommand\Subsubsection[2]{\subsubsection{#2}\label{sec:#1}}
\ifthenelse{\isundefined{\definition}}{\newtheorem{definition}{Definition}}{}
\ifthenelse{\isundefined{\assumption}}{\newtheorem{assumption}{Assumption}}{}
\ifthenelse{\isundefined{\proposition}}{\newtheorem{proposition}{Proposition}}{}
\ifthenelse{\isundefined{\theorem}}{\newtheorem{theorem}{Theorem}}{}
\ifthenelse{\isundefined{\lemma}}{\newtheorem{lemma}{Lemma}}{}
\ifthenelse{\isundefined{\corollary}}{\newtheorem{corollary}{Corollary}}{}
\ifthenelse{\isundefined{\alg}}{\newtheorem{alg}{Algorithm}}{}
\ifthenelse{\isundefined{\example}}{\newtheorem{example}{Example}}{}
\newcommand\cv{\ensuremath{\to}} \newcommand\cvL{\ensuremath{\xrightarrow{\mathcal{L}}}} \newcommand\cvd{\ensuremath{\xrightarrow{d}}} \newcommand\cvP{\ensuremath{\xrightarrow{P}}} \newcommand\cvas{\ensuremath{\xrightarrow{a.s.}}} \newcommand\eqdistrib{\ensuremath{\stackrel{d}{=}}} 

\newcommand\nl[1]{``\emph{#1}''}
\newcommand\wl[1]{\texttt{\footnotesize{#1}}}
\newcommand\score{\text{score}}
\newcommand\quantile{\text{quantile}}
\newcommand\single{{\small\textsc{Single}}}
\newcommand\comp{{\small\textsc{Comp}}}

\newcommand\Traverse{\mathbb{T}}
\newcommand\Member{\mathbb{M}}

\def\paren#1{\left( #1 \right)}
\newcommand{\ie}{{\it i.e.}}

\newcommand\pl[1]{\textcolor{red}{[PL: #1]}}

\title{Knowledge Representation in Graphs using Convolutional Neural Networks}

\author{Armando Vieira \\
Lidinwise, 1 Quality Court, WC2A 1HR London, UK\\
  {\tt armando@lidinwise.com} 
}
\date{}

\begin{document}

\maketitle

\begin{abstract}
Knowledge Graphs (KG) constitute a flexible representation of complex relationships between entities particularly useful for biomedical data. These KG, however, are very sparse with many missing edges (facts) and the visualisation of the mesh of interactions non-trivial.
Here we apply a compositional model to embed nodes and relationships into a vectorised semantic space to perform graph completion. A visualisation tool based on Convolutional Neural Networks and Self-Organised Maps (SOM) is proposed to extract high-level insights from the KG. We apply this technique to a subset of CTD, containing interactions of compounds with human genes / proteins and show that the performance is comparable to the one obtained by structural models. 
\end{abstract}

\section{Introduction}
Knowledge Graphs (KG) are knowledge representation structures commonly used to store complex structured or unstructured data.  Graphs can be direct or indirect with vertices representing entities (words, entities or concepts) and edges representing relationships between these entities. A KG contains two forms of knowledge: relational knowledge and categorical knowledge. Relational knowledge encodes the relationship between entities while categorical knowledge encodes the attributes of entities. 

Multi-relational graphs encodes data via knowledge databases, or semantic networks, and are widely
used in the Semantic Web and for knowledge representation in bioinformatics
(Gene Ontology, for instance) or natural language processing (WordNet).
In these graphs facts are modelled as triples in the form (subject, predicate, object), where a predicate either models the relationship between two entities or between an entity and an attribute value. Note that any information of the KB can be represented via a triple or a concatenation of several triples.
Such data sources are equivalently termed multi-relational graphs and they can be represented by a tensor,
where each component represents an adjacency matrix of a single predicate. 

Graph databases, such as Freebase \citep{bollacker2008freebase}, constitute rich repositories of annotated information that can be used for questions and answering applications. Simple or compositional questions can be formulated as, \nl{What chemical can increase expression of protein X in cell Y given that it was exposed to a substance Z?}.
However, KG are very incomplete and sparsely connected. An elegant solution to solve the data incompleteness is using vector space representations and control the dimensionality of the vectors to obtain good generalization on new facts
\citep{yang2015embeddings}. These methods are inspired in deep learning that is becoming the state-of-the-art approach for natural language processing tasks relying on distributed representations. 

For instance, the Word2vec model~\cite{mikolov}, have been proposed to capture the semantics of words through the context - the principle is that words that are semantically similar should be closer to similar context words.
The drawback of this approach is that the learned representations are mainly based on words, or entities, co-occurrences and cannot capture the relationship between two syntactically or semantically similar words if either of them yields very little context information.

In spite of their strong ability for representing complex data, multi-relational graphs are still complicated to understand, relations can be missing or invalid, there can be redundancy among entities because several nodes actually refer to the same concept, etc.
Furthermore, most multi-relational graphs are very large in terms of numbers of entities and of
relation types which make visualization knowledge representation hard - for example, Freebase contains more than 20 millions entities and billions of facts. 
Finally, most relations are localized on very few nodes - the so called fat-tail effect - making inference of new facts regarding most entities very hard.
 
Here we use a recent compositional methods for KG completion where the learning process is framed as the inference of new connections between nodes~\cite{compos}. This model addresses the problem of exploring relationships that goes beyond simple triples capturing chains of causality to generate and test hypothesis in the biomedical context. 

We also address the problem of visualisation of information contained in these graphs, a complex task as some nodes may have thousands of edges (relations) of dozens of different types. The combinatorial explosion of the number of possible causality relations severely constrains the use of graphical display tools. 

This work addresses these issues by applying a deep convolution network to visualize information contained in graphs with distributed representations. 
We demonstrate how to extract semantic fingerprints and show how they are useful for knowledge discovery and classification problems.

This paper is organised as follows: section 2 presents the model for embeddings and new facts discovery. Section 3 describes the data and Section 4 describes the SOM model, the semantic fingerprints obtained and some results. Section 5 describes the CNN model applied to the self-organized maps. Section 6 presents the an application to the protein compound prediction and Section 7 contains the conclusions and future work.




\section{The compositional model for graph embeddings}  \label{sec:comp_training}
Much work for knowledge graph completion was based on symbolic methods. 
These methods represent knowledge through simple algebraic operations but they are, in general, not tractable. 
Recently, a powerful approach for this task is to encode every element (entities and relations) of a
knowledge graph into a low-dimensional embedding vector space. 
Among these methods, TransE~\cite{bordes2014qa} is a simple and effective one with state-of-the-art link prediction precision. It learns low-dimensional embeddings for every entity and relation in the KG  where the basic idea is that every relation is regarded as a vectorial translation in the embedding space. 
For a triplet $(h, r, t)$, the embedding $h$ is close to the embedding $t$ by adding the embedding vector $r$,
so that $h + r \approx t$. TransE is suitable for 1-to-1 relations, but less robust dealing with 1-to-N,
N-to-1 or N-to-N relations.  
The model is training by minimising a loss function of how good new links are predicted in the test set.

Recently a compositional model was proposed~\cite{compos}. In this model the objective function to be minimised the following max-margin objective function:
\begin{equation}
	J(\Theta) = \sum_{i =1}^N \sum_{t' \in \mathcal{N}(q_i)} \left[1 - \text{margin}(q_i, t_i, t')\right]_{+}, \\
    \text{margin}(q, t, t') = \score(q, t) - \score(q, t'),
\end{equation}
where the parameters are the membership operator, the traversal operators, and
the entity vectors:
\begin{equation}
\Theta = \cup \left\{ T_{r}:r\in R\right\} \cup \left\{ x_{e} \in R^d : e\in E \right\}.
\end{equation}

The idea is to query not just simple triples but arbitrary complex ones from  a set of path query training examples  $\{(q_i, t_i)\}_{i=1}^N$ with path lengths ranging from 1 to $L$. While most models are trained only for objectives functions with queries of path length 1, this objective function takes into account extended composed paths.

\subsection{TransE model}
There are many possible implementations of $T$ and $M$, but we will use TransE~\cite{bordes2013translating} due to its simplicity and  performance on knowledge base completion.

Given a training set $S$ of triples $(h, l, t)$ composed of two entities $h$, $t\in \mathcal{E}$ (the set of entities) and a relationship $l\in L$ (the set of relationships), the model learns vector embeddings of the entities and the relationships that minimizes a given loss function. TransE works on the idea that a relation induced by the edges can be captured by a translation of the embeddings. 
It uses the scoring function:
\begin{align}
	\score(s/r, t) = -\| x_{s}  + w_{r} - x_{t} \|_2^2.
\end{align}
where $x_s$, $w_r$ and $x_t$ are all $d$-dimensional vectors and the model membership operator is defined as:
\begin{equation}
	(v, x_t) = -\|v - x_t \|_2^2
\end{equation}
and the traversal operator $T_r(x_s) = x_s + w_r$. TransE can handle a path query $q=s/r_1/r_2/\cdots/r_k$ using
\begin{equation}
	\score(q, t) = -\|x_s + w_{r_1} + \cdots + w_{r_k} - x_t \|_2^2.
\end{equation}

\subsection{Implementation}
We use Stochastic Gradient Descent (SGD)~\citep{duchi10adagrad} to optimize $J(\Theta)$, which is a non-convex objective function. We initialise all parameters with i.i.d. Gaussians of variance
0.25 and use a mini-batch size of 100 examples, and a step size in
$[0.001, 0.1]$ (chosen via cross-validation). 
For each example $q$, we sample 10 negative entities $t' \in \mathcal{N}(q)$. 
As suggested by ~\cite{compos}, during training, all of the entity vectors are constrained to lie on the unit ball, and we clipped the gradients to the median of the observed gradients if the
update exceeded 3 times the median. The dimensionality of the encodings was set to $d=50$, a margin of 1 and we use the L2 metric.

The models were implemented based on Theano libraries \citep{bastien2012theano} that are very fast and scalable.

%


\section{Data Description and Training}\label{sec:baseDataset}
\label{sec:compositionalize}
We will use two knowledge base completion datasets consisting of single-edge queries, a subset of WordNet~\cite{socher2013reasoning} and CTD~\cite{ctd}, as described in Table 1. The original CTD, contains around 575 150 nodes (genes, chemicals, diseases, proteins, etc) and 2 965 279 edges. For this work we only consider relations between compounds (chemicals) and genes - mostly, but not all, coding for proteins.
This information was extracted from manual annotations of around a hundred thousand scientific publications.

\begin{table}
\begin{centering}
\begin{tabular}{cc|c|c}
\multicolumn{2}{c|}{} & \textbf{WordNet} & \textbf{CTD}\tabularnewline
\hline
\multicolumn{2}{c|}{\textbf{Relations}} & 11 & 42 \tabularnewline
\multicolumn{2}{c|}{\textbf{Entities}} & 38 696 & 24 382\tabularnewline
\hline
 & Train & 112 581 & 316 321\tabularnewline
 & Test & 10 000 & 30 000\tabularnewline
\hline
\end{tabular}
\par\end{centering}
\protect\caption{\label{tab:datasets}Statistics of the databases used to train and test the models.}
\end{table}

CTD is very different from WordNet since it is a bipartite graph between compounds and genes and in WordNet both head and target entities are arbitrary words. In WordNet relations can be reversed but not in CTD. 
There are about 1700 different types of relationships between compounds and genes. We only consider the ones with more than 1000 facts, ending up with 42 relations types and 316 321 facts used for training.

Following~\cite{compos} we generate path queries by performing random walks on the graph thus generating an extended set of auxiliary training data. 

We start from the training graph $\mathcal{G}_{\mathrm{train}}$  formed by edges in the training set.
New training examples are generated as follows:
\begin{enumerate}
\item Uniformly sample a path length $L \in \{1, \dots, L_{\mathrm{max}}\}$, and uniformly sample
a starting entity $s \in \mathcal{E}$.
\item Perform a random walk beginning at entity $s$ and continuing $L$ steps.
\begin{enumerate}
\item At step $i$ of the walk, choose a relation $r_i$ uniformly from the set
of relations incident on the current entity $e$.
\item Choose the next entity uniformly from the set of entities reachable via $r_i$.
\end{enumerate}
\item Output a query-answer pair, $(q,t)$, where $q=s/r_1/\cdots/r_L$ and $t$ is the final entity of the random walk.
\end{enumerate}
Paths of length 1 were not sampled and we added all of the edges from $\mathcal{G}_{\mathrm{train}}$ to the dataset.

\section{The SOM visualisation} \label{sec:som}
Self-Organized Maps (SOM) is an algorithm for supervised or unsupervised clustering useful to represent high-dimensional data into a topographic two-dimensional map.
SOM were introduced by Kohonen~\cite{som} as an unsupervised learning process that learns the distribution of a set of patterns without any class information. A pattern is projected from an input space to a position in the map and information is coded as the location of an activated node. The advantage of SOM is that it provides a topological ordering of the classes where similarity is preserved and its useful for classification of data which a large number of categories - as is our case where is difficult to define class boundaries.
A SOM can be seen as a dimensionality reduction technique that use a neighbourhood function to preserve the topological properties of the input space.
 
A SOM operates in two modes: training and mapping. "Training" builds a map with input examples, while "mapping" classifies the unseen input vectors. In our case we use SOM to aggregate the embeddings obtained from the compositional TransE model. 

The advantage of SOM over other methods, like t-SNE, is that is simpler to interpret the results and it lends itself to a more flexible representation.
Next section explains how these maps can be interpreted and their usefulness in the biomedical context.

We used a rectangular 50x50 map with circular boundary conditions. Figure 2 describes the codevectors of the compounds embeddings for the CTD database using a embedding dimension of $d=50$ and the quality of each node.


\begin{figure}
\centering
\begin{minipage}{.5\textwidth}
  \centering
\includegraphics[scale=0.3]{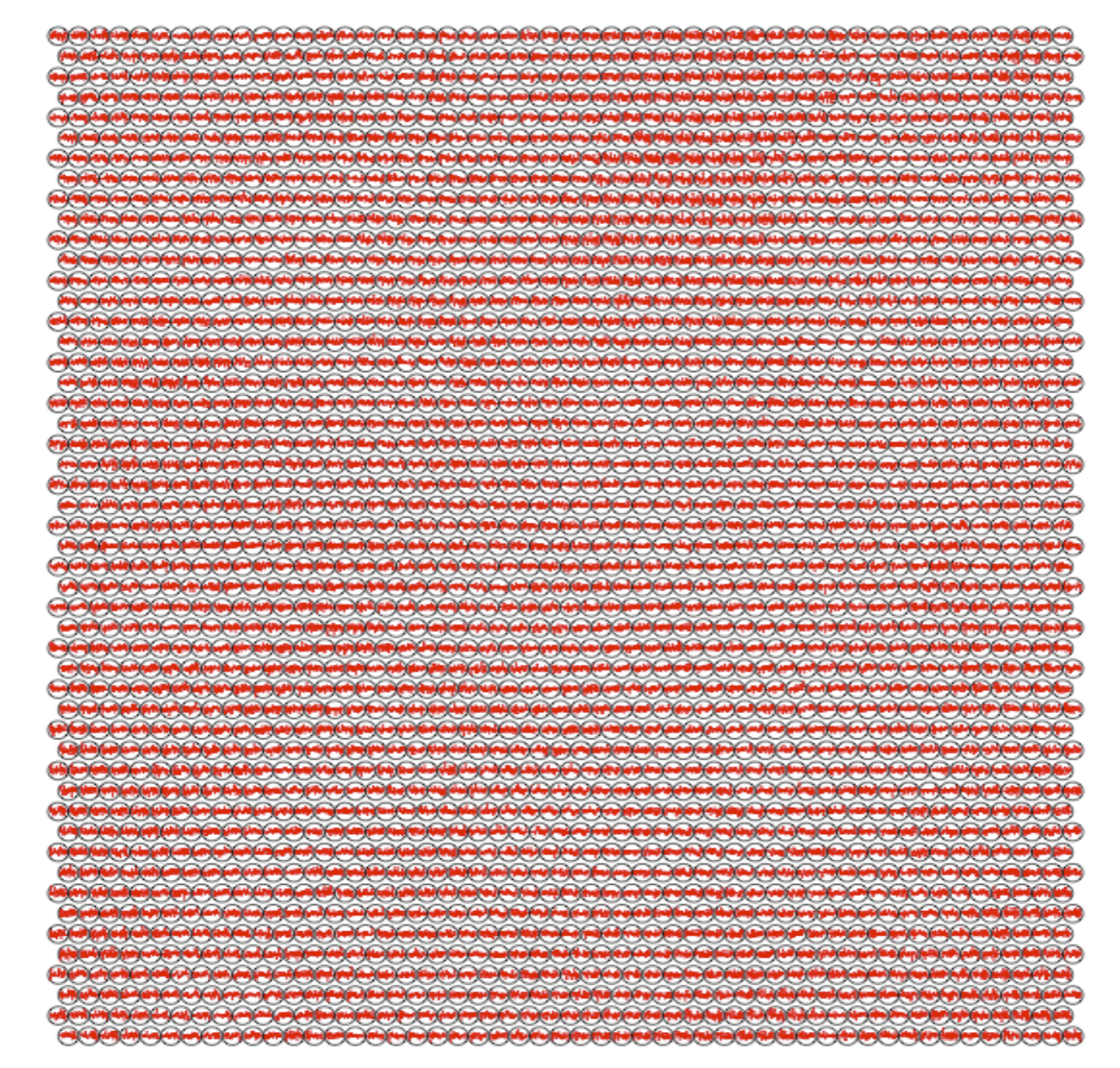}
  \label{fig:test1}
\end{minipage}%
\begin{minipage}{.5\textwidth}
  \centering
  \includegraphics[scale=.3]{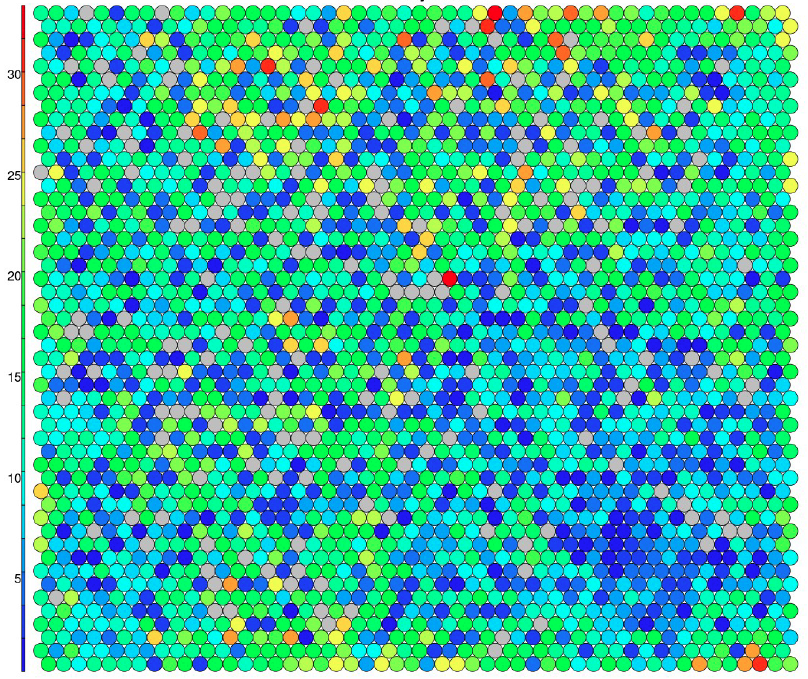}
  \label{fig:test2}
\end{minipage}
\protect\caption{\label{fig:som_code} Code-vectors obtained from the Self Organized Map for the embeddings of the CTD database and the quality of each node.}
\end{figure}


In Figure 2 we group the SOM cells into five categories, identified by colours. Each region correspond to a specific pattern of interaction between a set of genes and compounds. We can see that there are some well defined compact segments while others have a more scattered pattern (green). The explanation of these clusters will be detailed below.
 

\begin{figure}
\begin{centering}
\includegraphics[scale=0.75]{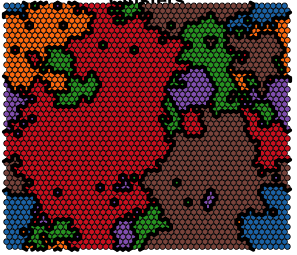}
\par\end{centering}
\protect\caption{\label{fig:som_gene} Self Organized Map of a gene-compound interaction pattern - here aggregated into 5 clusters.}
\end{figure}


In order to verify the results, we aggregate the SOM code-vectors into five clusters and project the interactions of the compounds with a set of genes aggregated by these clusters - see Table 2.

Ir order to access the quality of the projection, for a specific cell in the SOM map we select the chemicals associated with it, $C_{ij}$ and the genes that, in the TransE model, has a short distance. From the CTD database we them select the corresponding genes associated with these compounds and extract all the genes that have interactions with these chemicals. 
Based on these entities, we build a sub-graph and evaluate the Jaccard distances of all the chemical $C_{ij}$. 
We run an algorithm for all the cells in the SOM and compute a global ratio for evaluate the semantic capability of the clustering just obtained. The value was 1.37, which contrast with the initial value of 0.0031, a factor of about 400 - all type of interactions were considered. This shows that the aggregation made by the SOM has semantic meaning and is informative about the interactions in the original graph. 

\begin{table}
\begin{centering}
\begin{tabular}{lccccc}
Genes & \textbf{c1} & \textbf{c2}&  \textbf{c3}&  \textbf{c4} & \textbf{c5}\tabularnewline
\hline
All	 & 843 & 819 & 341 & 189 & 268 \tabularnewline
\hline
IL10	 & 3 & 42 & 59 & 103 & 5 \tabularnewline
\hline
EDN1	 & 2 & 23 & 51 & 119 & 1 \tabularnewline
\hline
UGT	 & 38 & 79 & 47 & 146 & 13 \tabularnewline
\hline
TIA	 & 1 & 0 & 1 & 49 & 0 \tabularnewline
\hline
AHR	 & 24 & 98 & 43 & 125 & 3 \tabularnewline
\hline
\end{tabular}
\par\end{centering}
\protect\caption{\label{tab:fingerprint}Results from the fingerprints in the gene space: number or compounds that interact with each gene and the cluster membership. }
\end{table}

For similar genes interaction patterns we expect similar mappings, as is fact the case for IL10 (interleukin 10) and EDN1 (endothelin 1). Note that this does not implies that the two genes are similar, only that they have a similar interaction pattern.


\subsection{From fingerprints to drug discovery}
Now that we know that the SOM represents meaningful information, we can go an extra step in terms of interpreting the results and help the researcher visualizing the data and test new hypothesis.

The traditional concept that drugs exert their activities by modulating one target of particular relevance to a disease has guided the pharmaceutical industry in recent decades.
However, there is evidence that this process is incomplete since many drugs do interact with multiple targets~\cite{ref7}.
Furthermore, a significant number of chemical compounds have failed to get approved due to serious clinical side-effects observed during later-stages [11]. 
The lesson in that multi-target interactions of drugs are largely unknown and poorly understood.

For a drug to have the desired effect we need to modulate a set of targets to achieve efficacy, while avoiding others to reduce the risk of side effects.
By considering all types of interactions between compounds and proteins, our method can be useful for the development multi target drugs.

Since we know the components involved in each disease, we can create its unique fingerprint based on the activation level of each code-vector in respect to the genes involved in the disease. In Figure 4 we plot the fingerprint for the lung and ovary cancer, i.e., genes that are involved in onset of the disease. We quantise the distances to the code-vectors into colours (red an euclidian distance below 0.1, green a distance between 0.1 and 0.2). All other higher distances were not consider.

The SOM fingerprint is helpful for visualizing the differences between the sets using distances between molecular fingerprints of the molecules. This technique clusters compounds (or genes) with similar interaction patterns with each other in the best matching cell while also maintaining a 2-dimensional grid of cells such that similar compounds or genes (depending on the mapping being used) appear in adjacent cells. Note that this pattern is not necessarily related to the structure or function of the genes/compounds being considered.




\begin{figure}
\centering
\begin{minipage}{.5\textwidth}
  \centering
  \includegraphics[scale=.5]{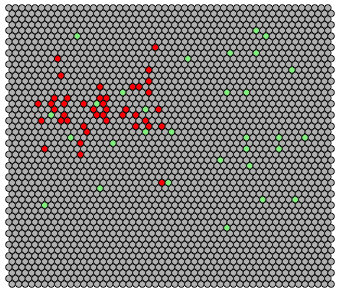}
  \label{fig:test1}
\end{minipage}%
\begin{minipage}{.5\textwidth}
  \centering
  \includegraphics[scale=.5]{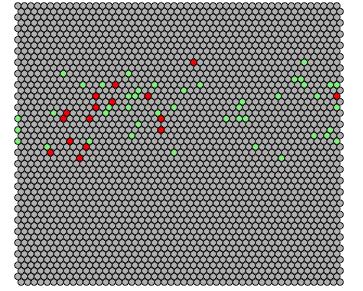}
  \label{fig:test2}
\end{minipage}
\protect\caption{ Fingerprint of a lung cancer and a ovary cancer}
\end{figure}

\section{Analysis of SOM maps with CNN}
SOM fingerprints are very useful for visualization but they are limited in terms of abstract features extraction.
In this section we will apply the findings learnt to build a supervised model for protein docking problem using a Convolutional Neural Network. CNN are powerful neuronal networks specially designed to capture invariant features in images setting the state-of-the-art in terms of image classification~\cite{cnn}. They also have a very interesting property in terms of capturing abstract representation of high-dimensional data. We will apply them to a very well known and important problem in structural biology: protein docking. We call this combination of SOM fingerprints and supervised CNN the SOME model.

This problem is in general ill-posed and not sufficiently constrained: many models can fit
the data thus achieving poor generalization.  Convolutional networks (CNN) incorporate hard constraints on learning and are good at detecting invariants on data, either to translation or deformation, which make them particularly useful for image analysis. They use three basic concepts: i) local receptive fields, ii) weight sharing, and iii) spatial subsampling.  

The network we will use consists of a set of layers, each of which contains one or more planes. Normalized images enter at the input layer and each unit receives input from a small neighborhood in the planes of the previous layer.  The weights forming the receptive field for a plane are forced to be equal at all points in the plane. Each plane can be considered as a feature map which has a fixed feature detector that is convolved with a local window which is scanned over the planes in the previous layer. Multiple planes are usually used in each layer so that multiple features can be detected. These layers are called convolutional layers. Once a feature has been detected, its exact location is less important. Hence, the convolutional layers are typically followed by another layer which does a local averaging and subsampling operation
(e.g., for a subsampling factor of two: where is the output of
a subsampling plane at position and is the output of the
same plane in the previous layer).  The network is trained with the usual Stochastic Gradient Descendent.

\subsection{SOME model}
\begin{figure}
\begin{centering}
\includegraphics[scale=.6]{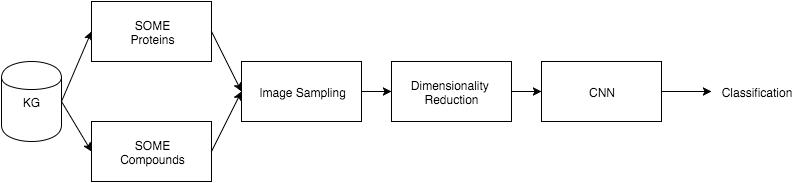}
\par\end{centering}
\protect\caption{\label{figsome} High level block model of the SOME+CNN model.}
\end{figure}

A high-level block diagram of the system proposed for protein docking prediction is shown in Fig~\ref{figsome} .

The system works as follows:

\begin{enumerate}
\item first we learn the genes and compounds embeddings from the CTD database using the compositional TransE model.

\item then we learn a SOM map that quantizes the $d$ dimensional input vectors into a sparse representation of 2500 topologically organized values - the fingerprints, or feature vectors. 

\item then, a fixed size window is used for the SOM map and local "image" samples are extracted. The window is moved over the map at each step.

 \item the training samples (compound + gene) are passed through the SOM at each step, thereby creating new training and test sets in the output space. 

\item Finally a convolutional neural network is trained on the transformed training/test set using supervised learning on the protein docking prediction. 

\end{enumerate}

For the SOM, training is split into two stages an ordering phase and a fine adjustment
phase. In the first phase 10 000 updates are executed, and 5 000 in the second. The learning rate during this phase is 0.5(1-$n/N$) where $n$ is the current update number, and $N$ the total number of updates. 


The idea of using CNN to analyse graphs embeddings projected into SOM maps is particularly interesting as it allows to extract abstract features from the high dimensional data, representing invariants that are hard to spot in other formats.

\section{Results} \label{sec:results}
We tested the compositional TransE model for link prediction and the SOME model for: i) consistency of clustering and ii) compound-target affinity prediction.

\subsection{Compositional TransE model}
In the first case we predict the fraction of new relations that are correctly predicted against a random set of random relations.  In Table 3 we present the results. The compositional model show considerable gain in respect to the single-node training, either for WordNet and for CTD. We used as metric the hits at top 10 (percentage of correct answers ranked in the top 10 predicted answers). On CTD improvement is even more remarkable on test set. 

\begin{table}[t]
\begin{centering}
\begin{tabular}{cc|cc|cc}
\multicolumn{2}{c|}{\textbf{Path query task}} & \multicolumn{2}{c|}{\textbf{WordNet}} & \multicolumn{2}{c}{\textbf{CTD}}\tabularnewline
\multicolumn{2}{c|}{} &@10& Class& @10 & Class\tabularnewline
\hline
\textbf{TransE} & $\single$ & 13.8 & 71.7 & 12.2 & 82.1\tabularnewline
 & $\comp$ & \textbf{43.5} & \textbf{87.4} & \textbf{27.4} & \textbf{87.2}\tabularnewline
\hline
 
\end{tabular}
\par\end{centering}
\protect\caption{\label{results}
Comparison of performance of single-edge training ($\single$) vs compositional training ($\comp$) for classification (Class) and top 10 hits prediction (@10).}
\end{table}

\subsection{SOME results}



Now we apply the SOME model to the prediction of protein ligand. This is a supervised learning problem where the objective is evaluate the ligand affinity of a specific compound to a protein. Normally this problem is addressed taking into consideration the structure of the molecule and the protein. 

For the SOME model we used a 50x50 grid where each cell is represented by the distance of the entities (head or tail) embeding to the the respective cell code-vectors. In this case we have two arrays: one for the proteins and one for the chemicals. 
For the chemicals we get an average of 2.2 compounds per cell and for the proteins/genes and average of 8.3 genes per SOM cell.

The CNN was trained using Keras framework (keras.io), a high level framework based in Theano libraries. The following configuration was used 
\begin{enumerate}
\item 8×24 input layer
\item convolutional layer with 71 3×3 filters with tanh activation
\item 2×2 max-pooling
\item convolutional layer with 88 2×2 filters with tanh activation
\item 2×2 max-pooling
\item a fully connected layer of size 26 with tanh activation
\item softmax classifier.
\end{enumerate}

The number of filters and the size of the fully connected layer were chosen using the method suggest by Snoek~\cite{snoek}. 
The cross entropy was used as the loss function and we used ReLU transfer function.. 
A dropout of 0.5. The convolutional network has six layers and for classification we use the softmax transformation. The network was trained with SGD for a total of 150 epochs.  As inputs of the CNN we used two SOM fingerprints maps; the protein and the compound.

We compare our model with a recent model called CSNAP~\cite{review} (Chemical Similarity Network Analysis Pulldown). This method address the problem that most target identification methods are limited to single ligand analyses. This method clusters diverse chemical structures into distinct sub-networks corresponding to chemotypes to improve target prediction accuracy achieving considerable improvement over traditional methods.

We tested our SOME model in the prediction of targets in a subset of annotated compounds. 
We used 100 ligands from 6 target-specific drug classes with known target annotations as a way to validate the method. As in~\cite{review} we used a chemical search criteria with a Z-score cutoff = 2.5 and a target class point of 0.85.  The overall prediction accuracy was 85\% which is slightly below the 89\% accuracy obtained by CSNAP. Note, however, that no information about chemical properties was used - just the embeddings. The performance of SOME could be improved if we enriched the inputs with contextual chemical information about the compounds, in the same line of~\cite{review}. 



 \section{Conclusions}
We proposed SOME, a graph completion and visualisation algorithm and applied it to biomedical data.  
SOME allows exploration of KG in a semantic meaningful representation and process queries that are impossible in traditional frameworks. For instance "Chemical$_ x$+Gene$_y$-Disease$D_1$+Disease$D_2$ = ?", or "Chemical$_x$ is to Gene$_y$ as Chemical$_{x1}$ is to ?".

Fingerprint matching is very useful to explore the high dimensional data since every entity can be projected in the global semantic space in the SOM map, thus producing an unique activation pattern. We can simply add or remove features (pixels) to one entity and see what is the implication in terms of relations other entities (diseases, for instance).

We showed that the visual exploration model proposed achieves comparative performance in protein docking problem using much less information and completely abstracting the chemical nature and composition of both elements.

As a future work we would like to explore the hierarchical clustering of SOM to allow the user to navigate through several levels of granularity when exploring the data. For the particular case of biomedical data, the inputs could be enriched with chemical context. We also would like to include full semantic context for diseases involving several genes so that the system may extract integrated approach.


\newpage
\bibliography{biblio}
\bibliographystyle{spr-chicago} 

\end{document}